# Short-Term Stock Price-Trend Prediction Using Meta-Learning

Shin-Hung Chang, Cheng-Wen Hsu, Hsing-Ying Li, Wei-Sheng Zeng and Jan-Ming Ho

*Abstract*— Although conventional machine learning algorithms have been widely adopted for stock-price predictions in recent years, the massive volume of specific labeled data required are not always available. In contrast, meta-learning technology uses relatively small amounts of training data, called fast learners. Such methods are beneficial under conditions of limited data availability, which often obtain for trend prediction based on time-series data limited by sparse information. In this study, we consider short-term stock price prediction using a meta-learning framework with several convolutional neural networks, including the temporal convolution network, fully convolutional network, and residual neural network. We propose a sliding time horizon to label stocks according to their predicted price trends, referred to as called slope-detection labeling, using prediction labels including "rise plus," "rise," "fall," and "fall plus". The effectiveness of the proposed meta-learning framework was evaluated by application to the S&P500. The experimental results show that the inclusion of the proposed meta-learning framework significantly improved both regular and balanced prediction accuracy and profitability.

I. INTRODUCTION

Time-ordered data have become widely available in recent years, and numerous time series classification (TSC) algorithms have been proposed. Owing to the natural complexity of time sequence information, some interpretation is almost always required to accurately model such data patterns. More specifically, for any classification problem on ordered time series data, some labeling consideration need to be kept in mind. The grouping of types of ordered data types can be understood as a TSC problem. At present TSC methods are commonly employed in many practical applications, including financial forecasting.

Machine learning models trained on historical financial datasets are increasingly used for stock price-trend predictions. Conventional time-series forecasting models are generally limited to the identification of linear relationships. As the result they are poorly suited to financial data, which is typically noisy, dynamic, and nonlinear [1]–[3]. Therefore, these models may not be ideal to represent the nonlinear relationships commonly encountered in the complex datasets relevant to stock price-trend prediction. Additionally, these conventional machine learning models are limited by their need for massive volumes of labeled data for training.

Stock price dataset generally include only 250 records per year or 2500 records per decade, for each stock. Furthermore, this sparse price dataset may not necessarily be representative of the dataset used to train the prediction model. This lack of training data in financial trend prediction is widely considered a significant challenge to the construction of well-tuned models. Thus, methods must be developed to process sparse training datasets which tend to lack examples for specific categories. In this study, we focus on meta-learning algorithms to solve this network model training problem using only relatively few samples.

The meta-learning framework requires a meta-learning network and several task learning mechanisms, which each classify a small number of samples [4]–[10]. The training process of the meta-learner is based on the principle of "learning to learn" which extends the scope of each task learner component from which knowledge can be derived. The meta-learner is presented with a number of tasks, on which two functions are performed. First, information is rapidly acquired from each of the tasks. Second, the acquired information from among the various tasks is processed at the higher level.

This study presents a novel meta-learning framework for stock price-trend prediction employing several classic neural network models, temporal convolution network (TCN) [13], fully convolutional network (FCN) [9] [11], and residual neural network (ResNet) [12]. We used a fixed time horizon, referred to as a slope-detection labeling method, to label stocks into four categories according to predicted price trends, including "rise plus," "rise," "fall," and "fall plus".

The effectiveness of the proposed meta-learning framework was evaluated by applying it to the Standard and Poor's 500 Index (S&P500). The remainder of this study is organized as follows. Section II summarizes the relevant literature. Section III presents our proposed meta-learning framework. Section IV details the experimental setup and results in terms of prediction accuracy, while concluding remarks and possible avenues for future research are given in Section V.

II. RELATED WORKS

The stock market has long been an important topic in finance and computation. The prediction of stock price trends poses unique challenges owing to the noisy data associated with the stochastic nature of financial data [2]. Many researchers have applied machine learning technologies to the prediction of stock prices, and have generally found that trading volumes had little impact on prediction performance when applied to the S&P500 and

Shin-Hung Chang is an assistant professor in Department of Computer Science and Information Engineering, Fu Jen Catholic University, 24205, New Taipei City, Taiwan. (phone:886-2-229053894; fax:886-2-29052442; e-mail: shchang@csie.fju.edu.tw).
Cheng-Wen Hsu is an undergraduate student in Department of Computer Science and Information Engineering, Fu Jen Catholic University, 24205, New Taipei City, Taiwan. (e-mail: chwnhsu@gmail.com).
Hsing-Ying Li is an undergraduate student in Department of Computer Science and Information Engineering, Fu Jen Catholic University, 24205, New Taipei City, Taiwan. (e-mail:xingyingli855@gmail.com).
Wei-Sheng Zeng is a research assistant in Institute of Information Science, Academia Sinica, 115, Taipei, Taiwan. (e-mail: wilson7126@gmail.com).
Jan-Ming Ho is a researcher in Institute of Information Science, Academia Sinica, 115, Taipei, Taiwan. (e-mail: hoho@iis.sinica.edu.tw).

DJIA datasets [14][16].

Such methods are generally categorized into two types, including individual and universal training models. Individual training models are trained using historical data from only single specific stock market [18]–[20]. In contrast, universal models are trained by feature extraction from a diverse of stock markets [23]. Regarding the former, Trafalis et al. [21] employed a support vector machine (SVM) to facilitate the prediction of prices over the short term, and Althelaya et al. [17] used long-term short-term memory (LSTM) to predict stock prices. Zhang et al. [15] used a hybrid model combining adaptive boosting (AdaBoost), a genetic algorithm (GA), and a probability SVM to predict the directions of stock price changes. However, the performance these individual models has thus far remained limited owing to the short-term training data employed. Among the latter, Fischer et al. [22] used an LSTM with large-scale data to predict the direction of price changes, and Hoseinzade et al. [23] adjusted a universal predictor using a variety of data sources to extract general characteristics of various stocks price changes.

Overall, universal models have achieved better prediction results than individual models. Additionally, convolutional neural models have demonstrated the ability to learn better by extracting important features to predict market patterns by using more highly diverse training datasets.

## III. PROPOSED META-LEARNING FRAMEWORK

In this paper, we propose a novel meta-learning framework for price trend forecasting based on financial time-series data. In contrast to previous studies, we applied this meta-learning framework to perform pre-training. These pre-training parameters were then used as the initialized parameter of the subsequent component learning models, which may be universal or individual models. The experimental results show that the proposed combined model improved prediction accuracy compared with the component models alone.

### A. Slope-detection Labeling Method

Labeling methods, Lunde [24] and Pagan [25], are often used to identify market trends, which are not sensitive to high frequency short-term stock price trends. We employed the proposed slope-detection method to define our stock price trends, including "rise plus," "rise," "fall," and "fall plus". In this paper, $p_d^n$ represents the closing price of stock $n$ on a target day $d$. $\bar{F}_d^n = \sum_{i=1}^{K} p_{d+i}^n$ denotes the average closing price of the stock $n$ in the next $K$ days after the day $d$. $\bar{B}_d^n = \sum_{i=-K+1}^{0} p_{d+i}^n$ relatively represents the average closing price of the stock $n$ in the past $K$ days, including the day $d$. Slope, $\delta_d^n$, is defined as $\bar{F}_d^n - \bar{B}_d^n$. $\mu_d^n$ denotes the mean value of the closing price and $\sigma_d^n$ is defined as standard deviation of the stock $n$ during these $2K$ days. The variable $\sigma_d^n$ was mathematically calculated using the following expression (1).

$$\sigma_d^n = \sqrt{\frac{\sum_{i=-K+1}^{K}(p_{d+i}^n - \mu_d^n)^2}{2K}} \quad (1)$$

We applied $\sigma_d^n$ as the threshold to label the price trend of detecting ascending (upward) or descending (downward) period according to slope, $\delta_d^n$. In the proposed method, if the closing price of the stock $n$ on the day $d$, $p_d^n$, is greater than $(\mu_d^n + \sigma_d^n)$ and $\delta_d^n > 0$, the trend of the stock $n$ on the day $d$ is labeled as "rise plus". If only condition $\delta_d^n > 0$ is satisfied, the trend is labeled as "rise". In contrast, if the closing price of stock $n$ on the day $d$, $p_d^n$, is less than $(\mu_d^n - \sigma_d^n)$ and $\delta_d^n < 0$, the trend of stock $n$ on the day $d$ is labeled as "fall plus". If only condition $\delta_d^n < 0$ is satisfied, the trend is labeled as "fall". In our proposed labeling method, "rise plus" indicates the end of a rising period (peak) and "fall plus" indicates the end of a falling period (trough). Additionally, "rise" means price trend is in an ascending period and "fall" means price trend is in a descending period. Table I presents the judgment conditions of our proposed slope-detection labeling method. Fig. 1 presents a labeling example of the proposed slope-detection method, where $K$ was set at 3.

TABLE I. JUDGEMENT OF SLOPE-DETECTION LABELING METHOD

| Trend Labeling | Judgement Condition |
|---|---|
| rise plus | $p_d^n > (\mu_d^n + \sigma_d^n)$ and $\delta_d^n > 0$ |
| rise | $\delta_d^n > 0$ |
| fall | $\delta_d^n < 0$ |
| fall plus | $p_d^n < (\mu_d^n - \sigma_d^n)$ and $\delta_d^n < 0$ |

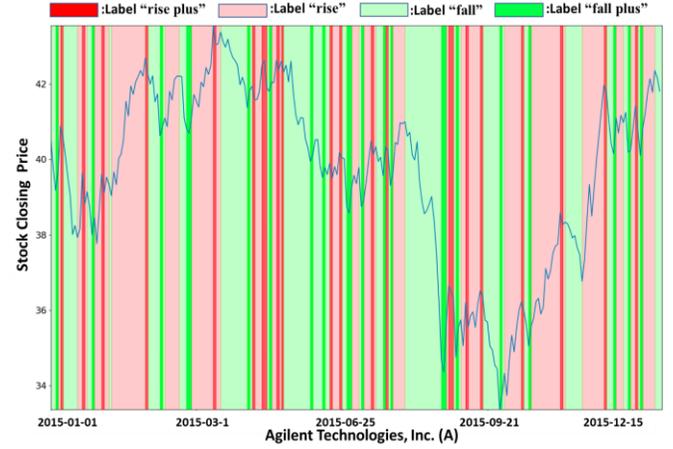

Figure 1. Labeling examples using slope-detection method between 2015-01-01 and 2015-12-31, where $K$=3.

### B. Designed Input Tensor

To increase profits, we selected our dataset from among various component stocks of the S&P500, which comprises well-established firms in the United States. The conventional approach employed by investors involves the observation of technical indicators based on the assumption that previous trading activity can provide important indicators of future trends. Accordingly, we selected the technical indicators presented in Table II, and used open-high-low-close (OHLC) prices as prediction features.

TABLE II. TECHNICAL INDICATORS

| Daily Price | |
|---|---|
| OHLC | Open-High-Low-Close Price |
| Technical indicators | |
| ATR | Average True Range |
| EMA20 | Exponential Moving Average |
| MOM6 | Momentum for 6 days |
| MOM12 | Momentum for 12 days |
| MA5 | Moving Average for 5 days |
| MA10 | Moving Average for 10 days |
| CCI | Commodity Channel Index |
| MACD | Moving Average Convergence Divergence |
| SMI | Stochastic Momentum Index |
| ROC | Rate of Change |
| WILLR | Williams %R |

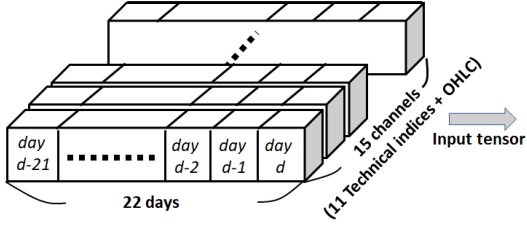

Figure 2. Structure of designed two-dimension input tensor.

Additionally, the proposed method utilizes an input tensor formed by combining these 11 technical indices and OHLC prices pertaining to each stock over a period of 22 days. In our short-term stock trend prediction model, we apply the selected 15 features of a stock over the previous 22 days, including day $d$, day $(d-1)$, day $(d-2)$, …, and day $(d-21)$, to predict the price trend at the day $d$. Therefore, we constructed a two-dimensional input tensor, as shown in Fig. 2.

We introduce the following notation for clarity. $\phi$ denotes the parameters of meta-learner, while $\theta_i$ indicates the parameters of each task-learner of stock $i$, $S_i$ represents the support set of stock $i$ for training each task-learner $\theta_i$ in the inner loop, $Q_i$ represents the query set for training a meta-learner $\phi$ in the outer loop, and α and β denote the learning rate used to update $\theta_i$ and $\phi$, respectively.

Our meta-learning framework was trained in two phases. The first phase, meta-training, involved training the meta-learner $\phi$ by evaluating gradients and computing adapted parameters. The second phase, meta-testing, involved fine-tuning all of the task-learners $\theta_i$ based on pre-trained $\phi$ trained using a training set $X_i$. We then evaluated the task learners using a testing dataset $E_i$.

The overall process is illustrated in Fig. 3. Stocks in different periods were defined as different tasks. In the meta-training phase, data from the year leading up to the previous month (training set) are used to train all of the tasks $\theta_i$, and then data from the previous month (testing set) are used to evaluate $\theta_i$. In the meta-testing phase, our objective was to ensure that the task learners performed well when applied to the testing dataset.

By applying the meta-training process, we obtain a general meta-learner $\phi$ including pre-trained parameters for all stocks of the S&P500. In the proposed method, the support set $S_i$ in one year is used to train each task-learner $\theta_i$ of stock $i$. After task learning, every query set $Q_i$ in one month is used to update meta-learner $\phi$, as presented in ① of Fig. 3.

Furthermore, the regular labeling of data from the previous period makes it possible to fine-tune the data from the current period and thereby update the meta-learner with time sliding to train the prediction capability of each task-learner $\theta_i$. The proposed meta-learning model also performs a fine-tuning process with horizontal time sliding mechanism to enable its. In one year, with a one-year training including each stock $X_i$, the model transfers the pre-trained meta-learner $\phi$ as a task-learner $\theta_i$ for each specific stock. This fine-tuning process of task-learner $\theta_i$ with the training set $X_i$ of each stock $i$ is processed based on one-month sliding, as presented in ② of Fig. 3.

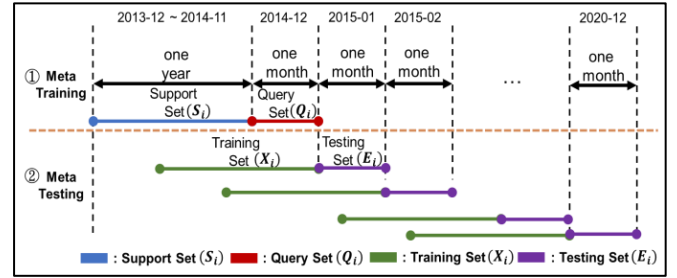

Figure 3. Proposed scenarios for meta training and meta testing.

### C. Meta-Training Algorithm

The meta-training process of our meta-learning framework is implemented in two phases. The first phase updates the parameters of each task-learner, which involves training task-learner $\theta_i$ of each stock $i$ with support set $S_i$. The support set $S_i$ is organized by sampling 20 records in each of the four labeling categories of stock $i$. Therefore, 40,000 records were randomly selected for 500 stocks of the S&P500, as shown in Fig. 4 (a). The second phase updates the parameters of meta-learner $\phi$. After five training epochs each task-learner $\theta_i$, query set $Q_i$ is input to each trained task-learner, and the summation of loss from each task-learner is calculated and applied to update meta-learner $\phi$, as shown in Fig. 4 (b). Additionally, in the first phase, we applied a sampled support set to avoid training failure caused by an unbalanced dataset, which influences the adaptability of the update process of the meta-learner.

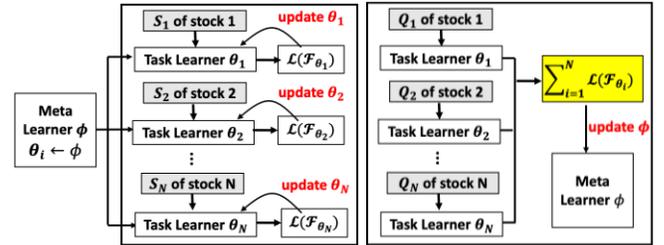

(a) Updating a task-learner $\theta_i$.　　(b) Updating a meta-learner, $\phi$.
Figure 4. Learning framework used to update task-learner and meta-learner.

| Algorithm 1: Meta-Learning algorithm for pre-training |
|---|
| Required: $\phi$: Meta learner; $\theta_i$: Task learner of each stock $i$;  $S_i$: Support set of stock $i$ ; $Q_i$: Query set of stock $i$;  α: Task learning rate; β: Meta learning rate; |
| 1. initialize $\phi$ |
| 2. initialize meta learner loss $\mathcal{L}_\phi$ |
| 3. **for** each iteration in **100** steps **do**　　/*Outer loop*/ |
| 4.　　**for** each stock $i$ with $S_i$ **do** |
| 5.　　　　$\theta_i \leftarrow \phi$ |
| 6.　　　　**for** each $S_i$ in **5** epochs **do**  /*Inner loop*/ |
| 7.　　　　　　update Task learner $\theta_i$ by gradient |
| 8.　　　　　　$\theta_i = \theta_i - \alpha \nabla_{\theta_i} \mathcal{L}(\mathcal{F}_{\theta_i})$ |
| 9.　　**for** each $\theta_i$ with $Q_i$ **do** |
| 10.　　　　$\mathcal{L}_\phi \leftarrow \mathcal{L}_\phi + \mathcal{L}(\mathcal{F}_{\theta_i})$ |
| 11.　　update meta learner $\phi$ by gradient |
| 12.　　$\phi = \phi - \beta \nabla_\phi \mathcal{L}_\phi$ |

The support set can be adapted to ensure that the latest features in each stock are used to train the model, such that $\theta_i$ fits each stock $i$. $\theta_i$ is updated via the gradient calculated using the following loss function (2).

$$\theta_i = \theta_i - \alpha \nabla_{\theta_i} \mathcal{L}(\mathcal{F}_{\theta_i}) \quad (2)$$

While training the task-learner, the number of epochs was usually kept low to prevent overfitting. Based on the approach adopted to learn overall stock trends, the meta-learner is updated in accordance with the loss summary for every stock in the query set, as shown in (3) below.

$$\mathcal{L}_\phi \leftarrow \mathcal{L}_\phi + \mathcal{L}(\mathcal{F}_{\theta_i}) \quad (3)$$

The difference between prediction and fine-tuning is that in the former, the query set does not update the meta-learner, rather, it only predicts the stock price trends. In formulating predictions for each subsequent period, we adaptively transform previous features of the meta-learner to be appropriate to the next time step as follows (4).

$$\phi = \phi - \beta \nabla_\phi \mathcal{L}_\phi \quad (4)$$

In preparing to predict stocks in real time, we do not set the initial $\theta$ randomly, but rather pre-train it using prior knowledge. The pre-training is based on the assumption that patterns closer to the target period may be expected to be more similar to data in the target period. Thus, our proposed method initially pre-trains the model and then continues fine-tuning it for every period up to the present. The overall process is described in Algorithm 1.

### D. Meta-Testing Algorithm for Fine Tuning

Two types of proposed meta-testing models are presented in this study, including individual and universal modes. In the individual model, the pre-trained meta-learner $\phi$ is transferred to each task-learner $\theta_i$, (as shown in Fig. 5). Pseudocode representing the steps of this individual model is presented in Algorithm 2. The following fine-tuning process is applied with each specific training set $X_i^t$ at one month $t$.

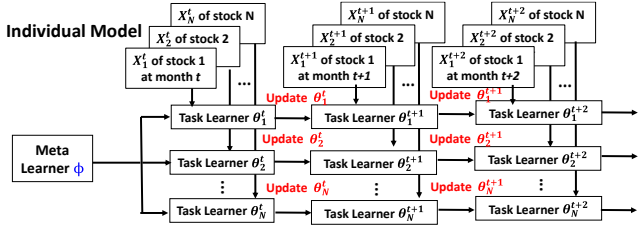

Figure 5. Meta-testing with individual task-learner.

**Algorithm 2: Individual fine tuning model algorithm**
Required: $X_i$: Training set of stock $i$; $E_i$: Testing set of stock $i$;
    $\phi$: Pre-trained meta-learner from Algorithm 1;
    $\theta_i$: Task learner of stock $i$;
    $\gamma$: fine trning learning rate;
1. $\theta_i \leftarrow \phi$
2. **for** each year with one month sliding **do**
3.      **for** each $X_i$ in each 50 epochs **do**
4.          update task learner $\theta_i$ by gradient
5.          $\theta_i \leftarrow \theta_i - \gamma \nabla_{\theta_i} \mathcal{L}(\mathcal{F}_{\theta_i})$
6.      **for** each stock $i$ with $E_i$ **do**
7.          Output prediction result from Task learner $\theta_i$

In contrast, in the universal mode, the pre-trained meta-learner $\phi$ is transferred only to a single task learner $\theta$. All the training sets $X^t$ are applied to fine-tune this task-learner $\theta$, as shown in Fig. 6. The process of this universal model is presented in detail in Algorithm 3.

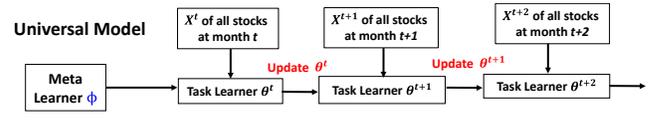

Figure 6. Meta-testing with universal task-learner.

**Algorithm 3: Universal fine tuning model algorithm**
Required: $X_i$: Training set of stock $i$; $E_i$: Testing set of stock $i$;
    $\phi$: Pre-trained meta-learner from Algorithm 1;
    $\theta$: Universal Task learner for all stocks;
    $\gamma$: fine trning learning rate;
1. $\theta \leftarrow \phi$
2. **for** each year with one month sliding **do**
3.      **for** each $X_i$ in each 50 epochs **do**
4.          update task learner $\theta$ by gradient
5.          $\theta \leftarrow \theta - \gamma \nabla_\theta \mathcal{L}(\mathcal{F}_\theta)$
6.      **for** each stock $i$ with $E_i$ **do**
7.          Output prediction results from universal Task learner $\theta$

## IV. EXPERIMENTAL RESULTS AND ANALYSIS

### A. Analysis of Proposed Labeling Method

#### 1) Two-level vs. Four-level labels

Most priors works have labeled price predictions as in a binary fashion as trend-up (rise) and trend-down (fall). However, in most investment situations, the two types of markers are often relatively weak indicators of the ground truth signals, and incorrect predictions are often obtained for values between small rises and small declines. Therefore, in this study, we hope that the proposed trend prediction model may achieve a high degree of investment credibility in making predictions. Therefore, we label stock price-trend predictions using four categories, including rise plus, rise, fall, and fall plus. The major goal of the four labeling categories is to predict the trend of a given stock, which is indeed a rising trend or falling trend.

#### 2) Threshold $\sigma$ of Slope-detection Method

In this study, we use the standard deviation $\sigma_d^n$ of the closing prices of stock $n$ from the past $K$ days to the next $K$ days as the threshold on the target day $d$. To observe the effectiveness of labeling the price trend in individual stocks, we compare the stock price $p_d^n$ of stock $n$ on the day $d$ with the this designed threshold $\sigma_d^n$.

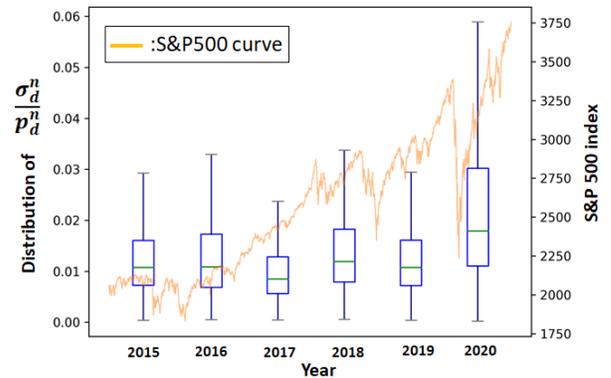

Figure 7. Box plot of $\frac{\sigma_d^n}{\mu_d^n}$ from year 2015 to year 2020.

In Fig. 7, taking the S&P500 index as an example, we observe that for more unstable market trends, the distribution of $\frac{\sigma_d^n}{p_d^n}$ was more dispersed, especially during the China-US trade war of 2018 and the COVID-19 pandemic of 2020. Therefore, our proposed slope-detection labeling method can indeed provide a clearer reflection of stock market trends. Furthermore, the experimental results demonstrate that our proposed labeling method was able to effectively tolerate drastic market changes.

*B. Parameter Settings of Experiments*

In our proposed meta-learning framework, we incorporated three different classical neural networks, include a fully convolutional network (FCN), a residual neural network (ResNet), and a temporal convolutional network (TCN). In addition, individual and universal models were implemented in our meta-testing process. Abbreviations for these model combinations are presented in Table III.

Table IV presents the hyperparameters used in our experiments. All our learning models used the "Adam" optimizer and the "CosineAnnealingLR" scheduling module in the training process. In the meta-training phase, we adjusted the task-learner for five epochs, and after every five adjustments, we adjusted the meta-learner once, and this training process was executed for a total of 100 generations. In the meta-testing phase, we adjusted the individual model and the universal model in 50 generations, and then selected the model hyper-parameters in the last generation.

TABLE III. MODEL ABBREVIATIONS.

| Models | NNs | Learning Architectures |
|---|---|---|
| Ind-NNs | FCN ResNet TCN | Individual + NNs |
| Uni-NNs | | Universal + NNs |
| Meta Ind-NNs | | Meta learning + individual + NNs |
| Meta Uni-NNs | | Meta learning + universal + NNs |

TABLE IV. PARAMETER SETTINGS.

| Learning Parameter | Value settings |
|---|---|
| α, β, γ | $1 \times 10^{-4}$ |
| Steps updating $\phi$ in Meta-Training | 100 |
| Epochs updating $\theta$ in Meta-Training | 5 |
| Epochs updating $\theta$ for universal model | 50 |
| Epochs updating $\theta$ for individual model | 50 |
| Epochs updating $\theta$ in Meta Testing | 50 |
| Optimizer | Adam |
| Scheduler | CosineAnnealingLR |

*C. Experimental Results*

Here, we evaluate the effectiveness of our proposed meta-learning framework in two aspects. First, we present the prediction accuracy of each model. Second, we address investment profitability based on the prediction results as an investment strategy from 2015-01-01 to 2020-12-31.

*1) Performance Metrics*

We address the stock trend prediction problem as a four-fold classification problem. The four labels used include "rise plus," "rise," "fall," and "fall plus". To avoid unfair experimental comparisons due to uneven labeling data, we applied regular accuracy, balance accuracy, and weighted F1-score to comprehensively evaluate the performance of our proposed models.

Table V shows that the accuracy of the universal models was better than that of the individual models by more than 10 percent. Additionally, models applying a meta-learning framework showed better performance in terms of prediction accuracy, balance accuracy, and weighted F1-score. ResNet with a meta-learning framework achieved the best "rise" prediction accuracy. For comparison, we evaluated the models' ability to distinguish between "rise" and "fall" by merging the labels "rise plus" and "rise" into "rise", and the labels "fall plus" and "fall" into "fall". Table VI shows the accuracy of only using these two levels of labels, "rise" and "fall". The best results were achieved by applying the meta-learning framework with the universal model.

TABLE V. PREDICTION ACCURACY WITH FOUR-LEVEL LABELS.

| Models | Regular Accuracy | Balance Accuracy | Weighted F1-Score | "rise" Precision |
|---|---|---|---|---|
| Ind-FCN | 45.19 % | 26.03 % | 39.54 % | 48.14 % |
| Meta-Ind-FCN | 48.59 % | 28.29 % | 44.29 % | 51.30 % |
| Ind-ResNet | 41.44 % | 25.98 % | 38.57 % | 47.73 % |
| Meta-Ind-ResNet | 45.68 % | 26.65 % | 41.43 % | 48.91 % |
| Ind-TCN | 36.76 % | 28.73 % | 39.45 % | 56.53 % |
| Meta-Ind-TCN | 37.04 % | 47.71 % | 39.14 % | 62.88 % |
| Uni-FCN | 59.82 % | 40.30 % | 57.37 % | 63.87 % |
| Meta-Uni-FCN | 59.33 % | 40.32 % | 57.02 % | 63.92 % |
| Uni-ResNet | 59.82 % | 40.30 % | 57.36 % | 64.09 % |
| Meta-Uni-ResNet | 60.09 % | 39.97 % | 57.50 % | 64.22 % |
| Uni-TCN | 62.06 % | 36.94 % | 57.31 % | 64.70 % |
| Meta-Uni-TCN | 62.14 % | 37.73 % | 57.81 % | 65.06 % |

TABLE VI. PREDICTION ACCURACY WITH TWO-LEVEL LABELS.

| Models | Regular Accuracy | Balance Accuracy | Weighted F1-Score |
|---|---|---|---|
| Ind-FCN | 54.38 % | 52.01 % | 51.37 % |
| Meta-Ind-FCN | 58.79 % | 57.52 % | 58.01 % |
| Ind-ResNet | 52.92 % | 51.52 % | 51.97 % |
| Meta-Ind-ResNet | 54.97 % | 53.37 % | 53.65 % |
| Ind-TCN | 60.21 % | 58.93 % | 59.47 % |
| Meta-Ind-TCN | 72.95 % | 73.29 % | 73.01 % |
| Uni-FCN | 76.31 % | 75.85 % | 76.24 % |
| Meta-Uni-FCN | 75.98 % | 75.64 % | 75.95 % |
| Uni-ResNet | 76.33 % | 75.94 % | 76.28 % |
| Meta-Uni-ResNet | 76.48 % | 76.10 % | 76.43 % |
| Uni-TCN | 77.42 % | 76.95 % | 77.35 % |
| Meta-Uni-TCN | 77.66 % | 77.25 % | 77.60 % |

*2) Investment Profitability Analysis*

We also compared various models in terms of investment profitability. Our stock trading strategy was to buy stocks held in S&P500 when the price trend signal of the stock was predicted to be "rise". Furthermore, these purchased stocks were sold if other three signals appeared on the next forecasting day with the price trend prediction model. In this study, we use the total cumulated return to show the investment profitability according to the prediction signal of different models and the proposed stock trading strategy. Additionally, we evenly distribute the funds (Equally Weighted Portfolio) to each stock that is predicted to have a signal of "rise". The value of our assets is calculated as the sum of the value of all stocks held which is calculated by multiplying number of each stock and the stock price on that day. In assessing profitability, we compared the cumulative return of different models from 2015-01-01 to 2020-12-31. As presented in Fig. 8, the investment profitability of models applying meta-learning frameworks with individual tuning model achieved better performance than those of the conventional models. In addition, we have observed that when applying a meta-learning framework with universal fine tuning models, profitability cannot be significantly improved. Different degrees of profit improvement are caused by the accuracy of the meta-learning framework for trend prediction.

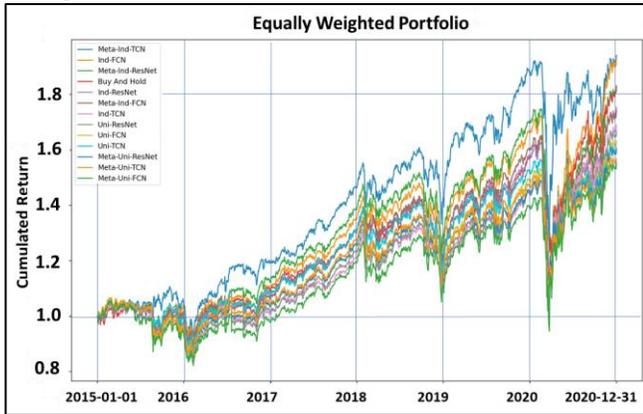

Figure 8. Cumulated return from 2015-01-01 to 2020-12-31.

## V. CONCLUSION

In this study, we have applied meta-learning to solve the time-series prediction problem for short-term stock price trends. We have proposed a novel slope-detection labeling method to divide the data into four categories, including "rise plus," "rise," "fall," and "fall plus," which demonstrated the ability to be more representative of stock market trends. We experimentally compared the effectiveness of different models on the basis of prediction accuracy and investment profitability. In terms of prediction accuracy, we focused on regular accuracy, balance accuracy, and weighted F1-scores. Our proposed meta-learning framework not only achieved higher accuracy than conventional methods, but also improves on their accuracy in avoiding the prediction of opposite trends. In terms of investment profitability, the Meta-Ind-TCN model was superior to the other models. Meta-Ind-TCN model could be trained satisfactorily to achieve greater than 70% accuracy with two-level labels. Furthermore, the cumulated return of using Meta-Ind-TCN model was able to grow more than 1.8 times the original investment amount. Thus, our proposed method demonstrated a significant improvement. In our future work, we will seek to improve forecast accuracy and profit performance further by considering other features, such as other technical indices and financial news.